\documentclass{article}
\usepackage{spconf,amsfonts,amsmath,graphicx,nicefrac,relsize}
\usepackage{graphicx}
\usepackage{subfig}
\usepackage{soul,color}
\usepackage{hyperref}
\title{PRRS Outbreak Prediction via Deep Switching Auto-Regressive Factorization Modeling }
\name{Mohammadsadegh Shamsabardeh$^{\dagger}$   \qquad Bahar Azari$^{\star}$ \qquad Beatriz Martínez-López$^{\dagger}$
\sthanks{This project was partially funded by the NSF BIGDATA:IA Award \#1838207 and NSF Track-D award \#2134901. Authors would like to acknowledge swine industry collaborators for the provision of data.}
}
  
\address{$^{\dagger}$Center for Animal Disease Modeling and Surveillance, University of California, Davis, CA, USA\\
$^{\star}$Northeastern University, Boston, MA, USA %\\$^{\ddagger}$Department of Veterinary Medicine and Epidemiology, University of California, Davis, USA
}

\begin{document}
%\ninept
%
\maketitle
\begin{abstract}
%We introduce an epidemic analysis framework for the prediction of PRRS virus outbreak that can capture the Spatio-temporal dynamics of infection transmission . 

We propose an epidemic analysis framework for the outbreak prediction in the livestock industry, focusing on the study of the most costly and viral infectious disease in the swine industry -- the PRRS virus. Using this framework, we can predict the PRRS outbreak in all farms of a swine production system by capturing the spatio-temporal dynamics of infection transmission based on the intra-farm pig-level virus transmission dynamics, and inter-farm pig shipment network. We simulate a PRRS infection epidemic based on the shipment network and the SEIR epidemic model using the statistics extracted from real data provided by the swine industry. We develop a hierarchical factorized deep generative model that approximates high dimensional data by a product between time-dependent weights and spatially dependent low dimensional factors to perform per farm time series prediction. The prediction results demonstrate the ability of the model in forecasting the virus spread progression with average error of NRMSE = $2.5\%$.

\end{abstract}


\section{Introduction}
\label{sec:intro}
\par The Porcine Reproductive and Respiratory Syndrome (PRRS) is arguably the most challenging and costly viral infectious disease in the US swine industry \cite{holtkamp2013assessment}. It is the high cost of disease control practices such as testing and vaccination as well as the financial damage caused by an outbreak that highlights the need to develop predictive models that can help to identify farms at high risk of infection to support risk-based, more cost-effective, target interventions. Such a framework will allow for more efficient testing, vaccination and outbreak prevention. Due to the high level of specialization in the swine industry, vast amount of data has been collected by the swine production systems. However, they have not yet been exploited enough due to difficulty of data access, integration and analyses (i.e., data are not consistently gathered, are non-standardized which makes their integration difficult, and are usually scattered across stakeholders). Examples of these data include diagnostic information, including the number of infected or dead animals, animal movements between farms or production data, which can give insights regarding farm health status and its contact network. Usually, these data do not satisfy the granularity required for learning an advanced predictive model (e.g. diagnostic samples are only taken once or twice per month per farm). However, using the real-world data, we can simulate epidemics to produce fine-grained time series data to analyze it further with an advanced novel prediction method based on a generative and variational inference model.  

\begin{figure}[t]
  \centering
  \subfloat[][farm-level]{
  \includegraphics[width=0.49\linewidth]{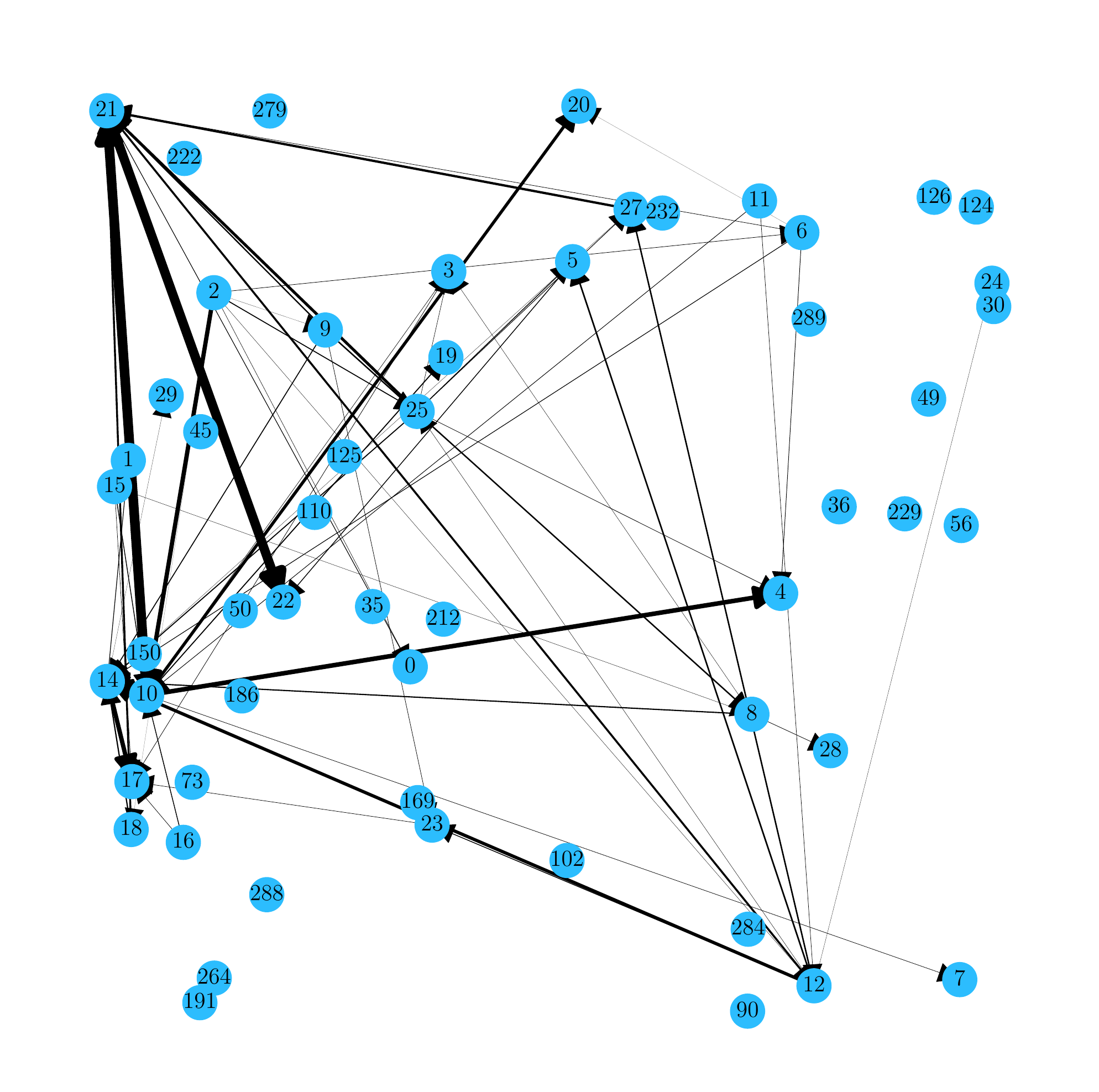}\label{fig:farmGraph}}
  \subfloat[][pig-level]{
  \includegraphics[width=0.49\linewidth]{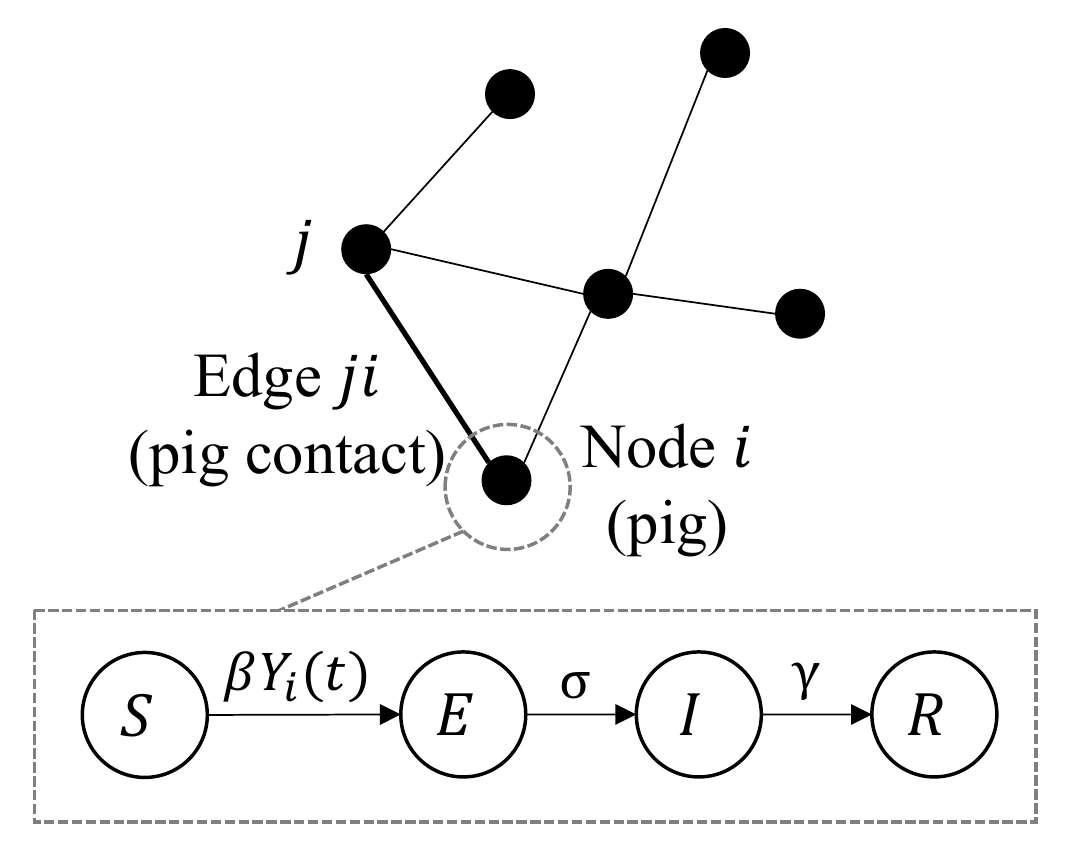}\label{fig:pigGraph}}
\caption{Contact Network. (a) The swine shipment network (directed graph). The premises are displayed by a number-labeled node and edge weights corresponds to the shipment rate. The between-premises shipment rate network is showcased for $10\%$ of nodes randomly selected among over 300 existing nodes.  (b) \textbf{Top:} Pig level network graph. \textbf{Bottom:} State-transition diagram for a single node.
\vspace{-.4cm}}
\label{fig:net}
\end{figure}

\par The direct contact with an infected pig is the main pathway for PRRS virus transmission. The network of between-farm movements, shown in Fig.~\ref{fig:farmGraph} together with the intra-farm (local) pig-level contact pattern based on Susceptible-Exposed-Infectious-Removed (SEIR) epidemic model (see Fig.~\ref{fig:pigGraph}) allows for constructing a system-level (global) pig-level disease transmission contact network, \cite{ferdousi2019generation}. Network-based SIR or SIR-extended epidemic models have been extensively studied in the literature \cite{lee2017unraveling}. In \cite{newman2002spread}, Newman studied a network-based SIR epidemic model where infection is transmitted through a random network of contacts between individuals. The disease transmission contact network is a probabilistic graph that can be sampled to generate virtual contact. Using this, we generate fine-grained spatio-temporal time-series data based on statistics of real-world data. %\vspace{-.5cm}
\par The spatio-temporal data is often considered to have a high level of correlation between spatial dimensions, and, therefore, they can be assumed to be governed by a smaller number of underlying components. For modeling the temporal dynamic of the time series including the number of infected, dead, or recovered pigs, we employ a non-linear vector auto-regressive latent model inspired by the work of \cite{farnoosh2021deep}. Our spatio-temporal time-series data is first \emph{factorized} into temporal weights and spatial factors. The temporal weights are modeled using a non-linear auto-regressive model parameterized by neural networks governed with a Markovian chain of discrete switches to capture higher-order multimodal latent dependencies, \cite{chang1978state, ghahramani1996switching, linderman2017bayesian, nassar2019tree,becker2019switching}.

\section{Time Series Data Simulation}
\label{sec:data}
\par Based on the rich database of an extensive anonymous swine production system located in the Midwest of the United States, we have access to farm-level pig shipment data, and PRRSV testing results \cite{shamsabardeh2019novel}. From 2006 to 2021, there have been over 260,000 movement records to or from farms within farm farm this production system. For each movement entry, the data includes the source and destination information, the number of transported pigs, and the date of the movement.  Based on the farm-level shipment data we generate a \emph{farm-level movement network} for the entire production system. Furthermore, the frequent PRRSV testing in each farm gives insight into how the virus is transmitted, e.g., what is the virus's transmission rate, incubation time, etc. Using the SEIR model we can produce a \emph{pig-level contact network}, \cite{ferdousi2019generation}. The combination of this two network built on the statistics extracted from our real data results in an intricate contact network by which we can simulate complex time series data showing the number of infected, dead, or recovered pigs in each farm in the production system.

\par To create the farm-level movement network we build a probabilistic graph, i.e., a graph in which the existence of edges is uncertain with some probability. A node in the graph represents a farm, while the weight of an edge is proportional to the shipment rate ( probability of shipment) between farms. Fig.~\ref{fig:farmGraph} shows a part of the shipment network of over 300 farms for our real data. The edge thickness is representative of the shipment rate. For local (intra-farm)  pig-level contact network in each farm, we consider a basic random graph model based on Erd\"{o}s—R\'{e}nyi model \cite{erre1959,ferdousi2019generation}, that produces pig contact graphs with an edge probability of $0.5$ between any pair of pigs. The global (inter-farm) pig-level contact network is constructed when we sample a random generalization of a between-farm shipment over time (each day) and as a result, we create pig contact between farms (see Fig.~\ref{fig:pigGraph}, top).

\par The simulation is formed on the network-based SEIR epidemic model for PRRS. In a network-based model, we consider a graph in which nodes represent individual pigs, and edges indicate direct or indirect contacts between pigs, which are considered infection pathways of PRRS. Each animal can be in one of the four states, Susceptible (S), Exposed (E), Infected (I), or Recovered (R) as the result of the epidemic progression. The state-transition diagram between these states is shown in figure.~\ref{fig:pigGraph}, Bottom. In the generated swine pig-level network, a PRRS outbreak is introduced by randomly selecting a pig farm, and infecting an arbitrary random number of pigs. We collect several time snapshots representing the progression dynamics of the disease spread, such as the number of infected pigs in each farm over time. The healthy pigs which are free from PRRS virus infection are classified as Susceptibles. If such a healthy pig comes into contact with infected pigs containing the virus, it may get infected at the rate $\beta Y_i(t)$, where $Y_i(t)$ is the number of infected neighbors of node $i$ at time $t$. If the transmission of pathogen occurs, a healthy pig enters into the Exposed group where it stays for the duration of the incubation period. On average, this period is denoted by $\nicefrac{1}{\sigma}$. Once it shows symptoms, it moves into the Infected group. It stays there for an average time of $\nicefrac{1}{\gamma}$ before it is recovered. We choose the parameters values $\beta = 0.087, \sigma = 7, \gamma = 6.5$ based on \cite{phoo2019modeling, charpin2012infectiousness}. This dynamic produces  a spatio-temporal time series from all farms over time that can be used for predication modeling in the next section (Sec.~\ref{sec:method}). 

%\vspace{-.5cm}

\section{Farm Disease Propagation Predication}
\label{sec:method}
\par Our spatio-temporal data indicates the number of pigs categorized within a particular stage, e.g., infected, recovered, etc., in every time instance in each farm. We denote this data as the matrix $X \in \mathbb{R}^{T \times D}$, where $T$ is the number of time points and $D$ the number of spatial locations, e.g., the number of farms. Building on previous work by \cite{farnoosh2021deep}, our assumption is that $X$ can be decomposed into a weighted summation of $K\ll D$ factors over time as: {\small
\begin{equation}
     X \approx [w_1,\cdots, w_T]^\top[f_1; \cdots; f_K] = W^\top F,
     \label{eqn:fac}
\end{equation}
}where $f_k\in\mathbb{R}^D$ is the $k^{\text{th}}$ spatial factor and $w_t\in\mathbb{R}^K$ is the weight vector at time $t$. Our intuition for adopting this model for some pig-specific collected measurements in $D$ farm over $T$ time points is that there are $K\ll D$ underlying factors using which we can approximate the overall dynamics of the disease propagation in the data. Our main goal is to predict the future outbreak behavior given the past time samples in each farm.

\begin{figure}[t]
    \centering
    \includegraphics[width=0.9\linewidth]{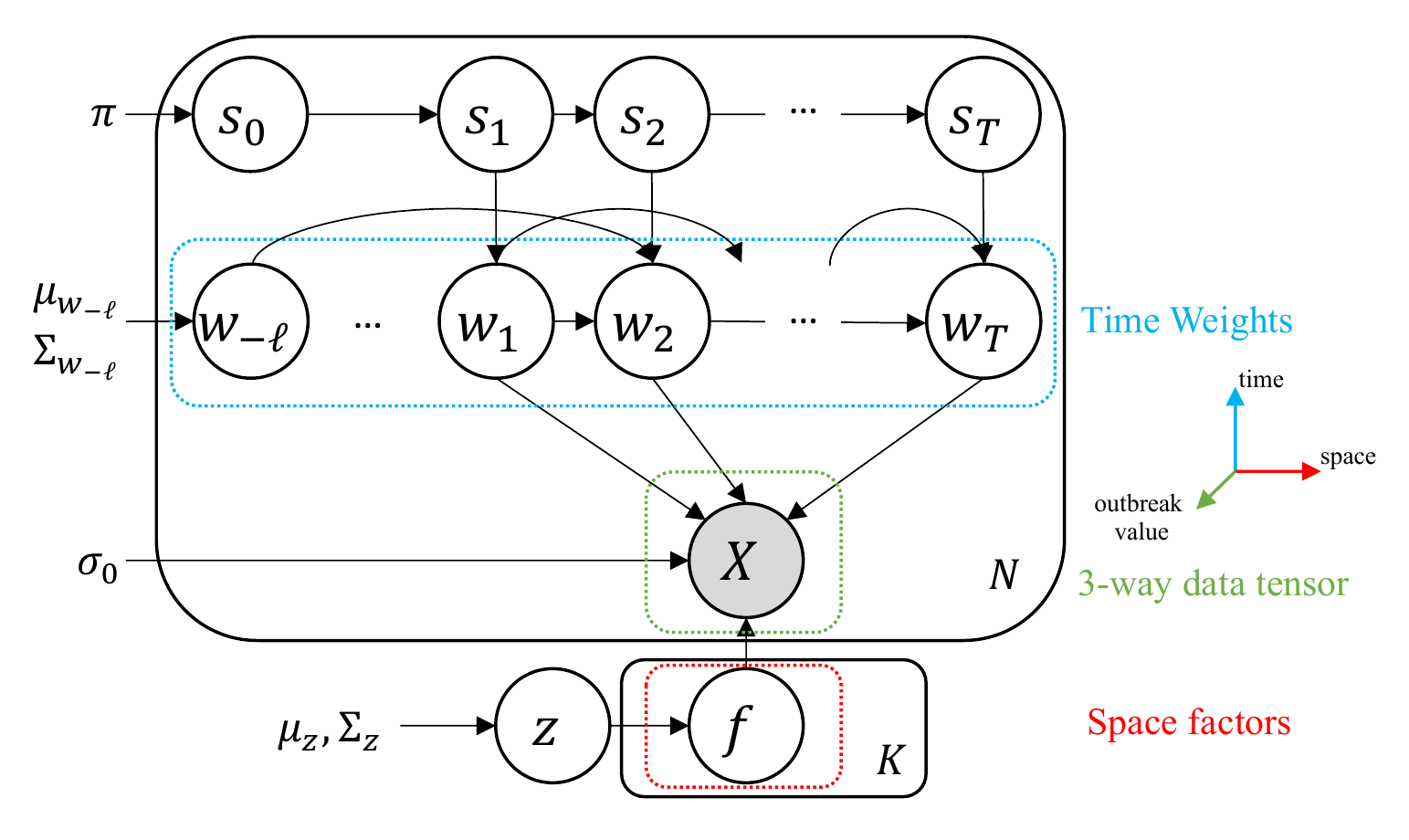}
    \caption{Probabilistic graphical model.}
    \label{fig:pgm}
\end{figure}

We assume that the weights, $W=\{w_{t}\}_{t=1}^T$, are generated according to a set of temporal lags, $\ell$, through a deep probabilistic switching auto-regressive model. These weights are furthermore governed by a Markovian chain of discrete latent states, $\mathcal{S}=\{s_{t}\}_{t=1}^T$ as follows: $w_t \sim p(w_t|w_{t-\ell},s_t),\; s_t \sim p(s_t|s_{t-1})$. In addition, we assume that spatial factors, $F=\{f_{k}\}_{k=1}^K$, are controlled by a shared low dimensional latent variable, $z$, as follows: $f_{1:K}\sim p(F|z),\;z\sim p(z)$. Fig.~\ref{fig:pgm} shows the relation among above-mentioned random variables in a probabilistic graphical model diagram form.

\begin{figure*}[!t]
\centerline{\includegraphics[width=\textwidth]{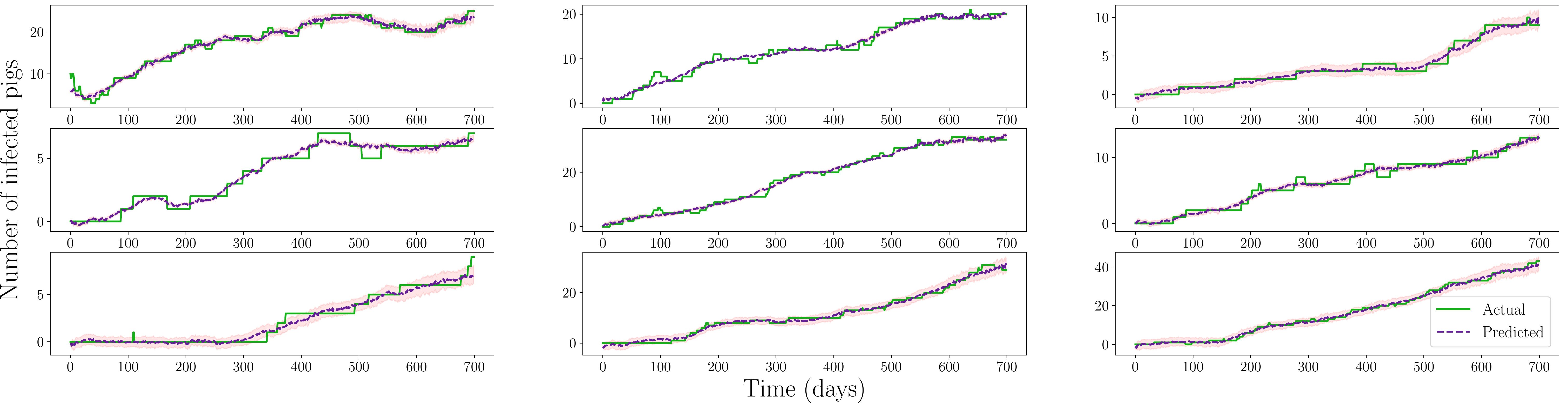}}
\caption{Short-term (one-day) predication. Each plot demonstrates the actual number of infected pigs in the simulated data (solid green), the mean estimate of the predictive model (dashed purple), and the standard deviation of the prediction estimate (shaded red error bar). Each row represents neighbouring farms that are connected in terms of pigs movement.}
\label{fig:res}
\end{figure*}

We train the model using stochastic variational methods \cite{hoffman2013stochastic,ranganath2013adaptive,kingma2013auto,rezende2015variational} by approximating the posterior $p_\theta(\mathcal{S}, W, z, F|X)$ using a variational distribution $q_\phi(\mathcal{S}, W, z, F)$, and by maximizing a lower bound (known as ELBO) $\mathcal{L}(\theta,\phi) \le \log p_\theta(X)$:

{\small
\begin{align}
    \mathcal{L}(\theta, \phi)
    &=
    \mathbb{E}_{q_\phi(\mathcal{S}, W, z, F)}
    \left[
    \log 
    \frac{p_\theta(X,\mathcal{S}, W, z, F)}
         {q_\phi(\mathcal{S}, W, z, F)}
    \right]\label{eqn:elbo}
    \\
    &=
    \log p_\theta(X)
    - 
    \text{KL}(q_\phi(\mathcal{S}, W, z, F) \,||\, p_\theta(\mathcal{S}, W, z, F|X)).\nonumber
\end{align}
%\graphicalmodel
}By maximizing the bound with respect to the parameters $\theta$, we learn the generative distribution over datasets $p_\theta(X)$, and by maximizing the bound over the parameters $\phi$, we do Bayesian inference by approximating the distribution $q_\phi(\mathcal{S}, W, z, F) \simeq p_\theta(\mathcal{S}, W, z, F|X)$ over latent variables for each data point. According to the graphical model in Fig.~\ref{fig:pgm}, the joint distribution of observations and latents will be:
{\small
\begin{align}
     \hspace{-.7em}p_\theta(X,\mathcal{S},\mathcal{Z}) = p(F|&z)p(z)\prod_{n=1}^N p(X_n|W_n, F) p(w_{n,-\ell}) p(s_{n,0})\nonumber\\
     &\prod_{t=1}^T p(s_{n,t}|s_{n,t-1}) p(w_{n,t}|w_{n,t-\ell},s_{n,t}),
     \label{eqn:pgen}
\end{align}} where $\mathcal{Z}=\{W, z, F\}$). Furthermore, we assume a fully factorized variational distribution for the latent variables posterior as:
{\small
\begin{flalign}
     \hspace{-.5em}q(\mathcal{S},\mathcal{Z}) = q(F)q(z)
     \prod_{n=1}^N q(w_{n,-\ell}) q(s_{n,0}) \prod_{t=1}^T
     q(s_{n,t}) q(w_{n,t}).\hspace{-.2em}
     \label{eqn:qinf}
\end{flalign}
}%\vspace{-1cm}
\par 
\subsection{Generative Parameters} 
Here we describe the generative distribution parameters and model assumptions. We assume the variable $s_{n,t}$ to represents a categorical variable of dimensionality $S$, i.e., the number of modes/switches that a system can be in at a specific time $t$. The sequence of the discrete latents, $s_{n,1:T}$, are in form of a Markov chain and govern the state transitions over time with distributions: {\small
\begin{align}
    p_\theta(s_{t}|s_{t-1}) &= \text{Cat}(\mathbf{\Phi}_\theta\,\mathlarger{\boldsymbol{\pi}}_{s_{t-1}})\nonumber\\
    q_\phi(s_{t-1}) &= \text{Cat}(\mathlarger{\boldsymbol{\pi}}_{s_{t-1}}),\label{eqn:state_transition}
\end{align}
}where $\mathlarger{\boldsymbol{\pi}}_{s_{t-1}} = [\pi_1, \cdots, \pi_S]$ represents the probabilities of the categorical distribution for $s_{t-1}$, and $\mathbf{\Phi}_\theta\in\mathbb{R}^{S\times S}$ is a valid probability transition matrix.

\par For the temporal weights, $w_{t}$, we assume a switching Gaussian dynamic for the temporal latent transitions governed by the discrete latent states, $s_t$. In other words, we assume that the marginal distribution of temporal weights follows a Gaussian mixture distribution in the latent space, as:
{\small
\begin{equation}
   p_\theta(w_{t}|w_{t-\ell},s_{t} = s) = \mathcal{N}\Big(\mu_{\theta_s}^w(w_{t-\ell}), \Sigma_{\theta_s}^w(w_{t-\ell})\Big),\nonumber
\end{equation}
}where $s\in\{1,\cdots,S\}$, and state-specific $\mu_{\theta_s}^w(\cdot)$ and diagonal $\Sigma_{\theta_s}^w(\cdot)$ are parameterized by multilayer perceptrons (MLPs), hence, follow a \emph{nonlinear} vector auto-regressive model given $w_{t-\ell}$. Namely, we feed $w_{t-\ell}$ to a multi-head MLP for estimating the Gaussian parameters, e.g.,
{\small
\begin{align}
\mu_{\theta_s}^w=\text{FC}_s(h_s),\quad
h_s = \sum_{l\in\ell}\sigma(\text{FC}_{s,l}(w_{t-l})),\nonumber
\end{align}
}where FC denotes a fully connected layer, and $\sigma$ is a non-linear activation function. 

\par For the spatial factors, $F$, we assume a diagonal Gaussian distribution for spatial factors parameterized with an MLP as 
{\small
\begin{equation}
p_\theta(F|z) = \mathcal{N}\big(\mu_\theta^F(z), \Sigma_\theta^F(z)\big)\end{equation}
}where $z$ is sampled from a normal distribution: $z \sim \mathcal{N}(0, I)$. The latent $z$ is introduced as a low dimensional spatial embedding that encourages the estimation of a multimodal distribution among spatial factors. Given the temporal weights and spatial factors, we reconstruct the data by consolidating the two factorized part as:
{\small
\begin{equation}
    X_n \sim p_\theta(X_n|W_n,F) =
    \mathcal{N}\Big(\big[w_{n,1}, \cdots, w_{n,T}\big]^\top F, \sigma_0^2\Big),%\nonumber
\end{equation}
}where $\sigma_0$ is a hyperparameter for observation noise.

\subsection{Variational Parameters}
The trainable variational parameters, $\phi$, are assumed to have fully factorized distributions. The variational distribution for the continuous variables, $q(z;\phi^{z})$, $q(F;\phi^F)$, and $\{q(w_{n,t};\phi^w_{n,t})\}_{n=1,\,t=-\ell}^{N,\,T}$, are considered to be Gaussian distributions with diagonal covariances. In addition, the variational parameters for the distribution of discrete latents,  $\big\{q(s_{n,t};\phi^s_{n,t})\big\}_{n=1,\,t=1}^{N,\,T}$, are considered based on the mean-field approximation assumption to compensate information loss (see \cite{farnoosh2021deep} for more detail.)

\subsection{Training Procedure}
The Monte-Carlo estimate of the gradient of ELBO is computed with respect to generative, $\theta$, and variational, $\phi$, parameters using a re-parameterized sample, \cite{kingma2013auto}, from the posterior of continuous latents, $\{W, z, F\}$. For the discrete latent, $\mathcal{S}$, however, we compute the expectations by summing over the $S$, without the need for explicit sampling. This regularizes the $S$ nonlinear auto-regressive priors based on
their corresponding weighting.
We can analytically calculate the Kullback-Leibler (KL) divergence terms of ELBO for both multivariate Gaussian and categorical distributions, which leads to lower variance gradient estimates and faster training as compared to e.g., noisy Monte Carlo estimates often used in literature. We use the Adam optimizer, \cite{kingma2014adam}, with learning rate of $0.01$ for training. We initialized all the parameters randomly, and adopted a linear KL annealing \cite{bowman2015generating} schedule to increase from $0.01$ to $1$ over the course of $100$ epochs.

\section{Experimental Results}
\label{sec:results}
We used time series of epidemic progression from over 300 farms simulated for 700 time points. We kept last $20\%$ of the time series as the test set.  We then performed a short-term prediction tasks by adopting a rolling prediction scheme reported in \cite{chen2019missing}. For short-term prediction, the next time point is predicted on the test set using the generative model and spatial factors learned on the train set. We reported the test set normalized root-mean-square error (NRMSE\%), which is related to the expected negative test-set log-likelihood for the case of Gaussian distributions, and it is used for evaluating the predictive generative models. We obtained NRMSE of $2.5\%$ averaged over all the farms. Fig.~\ref{fig:res} shows the number of infected pigs over time for nine selected farms. In this figure, we illustrate the actual number of infected pigs in the simulated data with a solid green curve, the mean estimate of the predictive model with a dashed purple curve, and the standard deviation of the data with a shaded red error bar. Each row in the figure represents a group of relatively highly connected farms in terms of the frequency of the pig shipments. Note that each group show a relatively strong correlation regarding the outbreak progression and the predictive model was able to capture this.    

One observation regarding the performance of the model in situations when we do not have a curve with smooth behavior is the fact that when the increase and decrease of the number of infected pigs is abrupt, the model tends to converge to a middle point between the current value and the future one. This issue can be addressed using a model selection choice. Specifically, we can control the number of factors $K$ in order to increase the flexibility of the model in capturing different levels of curve smoothness.    
\section{Discussion \& Conclusion}
\label{sec:conc}

The PRRS outbreak cause an economic loss of over \$664 million annually \cite{holtkamp2013assessment}, which can be significantly mitigated by early detection and risk-based intervention practices. Direct contact is the main disease transmission pathway. Therefore, the pig contact network provides a substantial basis to develop an outbreak prediction framework. We create a system-wide pig contact network by combining the SEIR epidemic model based on intra-farm infection transmission parameters, and inter-farm pig shipment network. We presented a hierarchical factorized deep generative model of our spatio-temporal data that can capture the underlying dynamics of the disease spread with the aim to predict the number of infected pigs in all farms. Our result demonstrates the ability of the model in forecasting the virus spread progression with an average one-day prediction error of NRMSE = $2.5\%$.  A potential future direction is to incorporate the per farm disease transmission parameters for the SEIR model to represent variations in the disease control implementation. Additionally, indirect disease transmission pathways, such as airborne, can be included in the framework for a more comprehensive analysis.

%\vfill
%\pagebreak

%\newpage
%\vfill\pagebreak

\bibliographystyle{IEEEbib}
\bibliography{ICASSP2022}

\begin{thebibliography}{10}

\bibitem{holtkamp2013assessment}
Derald~J Holtkamp, James~B Kliebenstein, Eric~J Neumann, Jeffrey~J Zimmerman,
  Hans~F Rotto, Tiffany~K Yoder, Chong Wang, Paul~E Yeske, Christine~L Mowrer,
  and Charles~A Haley,
\newblock ``Assessment of the economic impact of porcine reproductive and
  respiratory syndrome virus on united states pork producers,''
\newblock {\em Journal of Swine Health and Production}, vol. 21, no. 2, pp.
  72--84, 2013.

\bibitem{ferdousi2019generation}
Tanvir Ferdousi, Sifat~Afroj Moon, Adrian Self, and Caterina Scoglio,
\newblock ``Generation of swine movement network and analysis of efficient
  mitigation strategies for african swine fever virus,''
\newblock {\em PloS one}, vol. 14, no. 12, pp. e0225785, 2019.

\bibitem{lee2017unraveling}
Kyuyoung Lee, Dale Polson, Erin Lowe, Rodger Main, Derald Holtkamp, and Beatriz
  Mart{\'\i}nez-L{\'o}pez,
\newblock ``Unraveling the contact patterns and network structure of pig
  shipments in the united states and its association with porcine reproductive
  and respiratory syndrome virus (prrsv) outbreaks,''
\newblock {\em Preventive veterinary medicine}, vol. 138, pp. 113--123, 2017.

\bibitem{newman2002spread}
Mark~EJ Newman,
\newblock ``Spread of epidemic disease on networks,''
\newblock {\em Physical review E}, vol. 66, no. 1, pp. 016128, 2002.

\bibitem{farnoosh2021deep}
Amirreza Farnoosh, Bahar Azari, and Sarah Ostadabbas,
\newblock ``Deep switching auto-regressive factorization: Application to time
  series forecasting,''
\newblock in {\em Proceedings of the AAAI Conference on Artificial
  Intelligence}, 2021, vol.~35, pp. 7394--7403.

\bibitem{chang1978state}
Chaw-Bing Chang and Michael Athans,
\newblock ``State estimation for discrete systems with switching parameters,''
\newblock {\em IEEE Transactions on Aerospace and Electronic Systems}, , no. 3,
  pp. 418--425, 1978.

\bibitem{ghahramani1996switching}
Zoubin Ghahramani and Geoffrey~E Hinton,
\newblock ``Switching state-space models,''
\newblock Tech. {R}ep., Citeseer, 1996.

\bibitem{linderman2017bayesian}
Scott Linderman, Matthew Johnson, Andrew Miller, Ryan Adams, David Blei, and
  Liam Paninski,
\newblock ``Bayesian learning and inference in recurrent switching linear
  dynamical systems,''
\newblock in {\em Artificial Intelligence and Statistics}, 2017, pp. 914--922.

\bibitem{nassar2019tree}
J~Nassar, SW~Linderman, M~Bugallo, and IM~Park,
\newblock ``Tree-structured recurrent switching linear dynamical systems for
  multi-scale modeling,''
\newblock in {\em International Conference on Learning Representations (ICLR)},
  2019.

\bibitem{becker2019switching}
Philip Becker-Ehmck, Jan Peters, and Patrick van~der Smagt,
\newblock ``Switching linear dynamics for variational bayes filtering,''
\newblock 2019.

\bibitem{shamsabardeh2019novel}
M~Shamsabardeh, Shabaz Rezaei, Jose~Pablo Gomez, Beatriz
  Mart{\'\i}nez-L{\'o}pez, and Xin Liu,
\newblock ``A novel way to predict prrs outbreaks in the swine industry using
  multiple spatio-temporal features and machine learning approaches,''
\newblock {\em Frontiers in Veterinary Science}, vol. 6, 2019.

\bibitem{erre1959}
P.~Erdos and A.~Rényi,
\newblock ``On random graphs i,''
\newblock {\em Publicationes Mathematicae (Debrecen)}, vol. 6, pp. 290--297,
  1959.

\bibitem{phoo2019modeling}
Phithakdet Phoo-ngurn, Chanakarn Kiataramkul, and Farida Chamchod,
\newblock ``Modeling the spread of porcine reproductive and respiratory
  syndrome virus (prrsv) in a swine population: transmission dynamics, immunity
  information, and optimal control strategies,''
\newblock {\em Advances in Difference Equations}, vol. 2019, no. 1, pp. 1--12,
  2019.

\bibitem{charpin2012infectiousness}
C{\'e}line Charpin, Sophie Mah{\'e}, Andr{\'e} Keranflec’h, Catherine Belloc,
  Roland Cariolet, Marie-Fr{\'e}d{\'e}rique Le~Potier, and Nicolas Rose,
\newblock ``Infectiousness of pigs infected by the porcine reproductive and
  respiratory syndrome virus (prrsv) is time-dependent,''
\newblock {\em Veterinary Research}, vol. 43, no. 1, pp. 1--11, 2012.

\bibitem{hoffman2013stochastic}
Matthew~D Hoffman, David~M Blei, Chong Wang, and John Paisley,
\newblock ``Stochastic variational inference,''
\newblock {\em The Journal of Machine Learning Research}, vol. 14, no. 1, pp.
  1303--1347, 2013.

\bibitem{ranganath2013adaptive}
Rajesh Ranganath, Chong Wang, Blei David, and Eric Xing,
\newblock ``An adaptive learning rate for stochastic variational inference,''
\newblock in {\em International Conference on Machine Learning}, 2013, pp.
  298--306.

\bibitem{kingma2013auto}
Diederik~P Kingma and Max Welling,
\newblock ``Auto-encoding variational bayes,''
\newblock {\em stat}, vol. 1050, pp. 1, 2014.

\bibitem{rezende2015variational}
Danilo~Jimenez Rezende and Shakir Mohamed,
\newblock ``Variational inference with normalizing flows,''
\newblock in {\em International Conference on Machine Learning}, 2015, pp.
  1530--1538.

\bibitem{kingma2014adam}
Diederik~P Kingma and Jimmy Ba,
\newblock ``Adam: A method for stochastic optimization,''
\newblock {\em arXiv preprint arXiv:1412.6980}, 2014.

\bibitem{bowman2015generating}
Samuel~R Bowman, Luke Vilnis, Oriol Vinyals, Andrew~M Dai, Rafal Jozefowicz,
  and Samy Bengio,
\newblock ``Generating sentences from a continuous space,''
\newblock {\em CoNLL 2016}, p.~10, 2016.

\bibitem{chen2019missing}
Xinyu Chen, Zhaocheng He, Yixian Chen, Yuhuan Lu, and Jiawei Wang,
\newblock ``Missing traffic data imputation and pattern discovery with a
  bayesian augmented tensor factorization model,''
\newblock {\em Transportation Research Part C: Emerging Technologies}, vol.
  104, pp. 66--77, 2019.

\end{thebibliography}

\end{document}